\documentclass[5p,times]{elsarticle}

\usepackage{lineno,hyperref}
\usepackage{graphicx}

\usepackage[Symbol]{upgreek}
\usepackage{amssymb}

\usepackage{pifont}
\usepackage[utf8]{inputenc}
\usepackage{caption}
\usepackage{subcaption}
\usepackage[english]{babel}
\usepackage{graphicx}
\usepackage{amsmath}
\usepackage{lscape}
\usepackage{amssymb}
\usepackage{esvect}
\usepackage{multirow}
\usepackage{footmisc}

\usepackage{mathtools}
\usepackage{url}
\usepackage{mathtools, cuted}
\usepackage{algorithm}
\usepackage{algpseudocode}
\usepackage{boxhandler}
\usepackage{verbatim}
\usepackage{blindtext}
\usepackage{xcolor}
\usepackage{caption}
\usepackage{subcaption}
\usepackage{adjustbox}

\journal{Journal of Network and Computer Applications}

\begin{document}
\begin{frontmatter}

\title{\textbf{STOPPAGE}:  \textbf{S}patio-\textbf{t}emporal Data Driven Cl\textbf{o}ud-Fog-Edge Com\textbf{p}uting Framework for \textbf{Pa}ndemic Monitoring and Mana\textbf{ge}ment}

\author[mymainaddress]{Shreya Ghosh}
\ead{shreya.cst@gmail.com}
\author[mymainaddress1]{Anwesha Mukherjee}
\ead{anweshamukherjee2011@gmail.com}

\author[mymainaddress]{Soumya K Ghosh}
\ead{skg@cse.iitkgp.ac.in}

\author[mysecondaryaddress]{Rajkumar Buyya}
\ead{rbuyya@unimelb.edu.au}

\address[mymainaddress]{Department of Computer Science and Engineering, Indian Institute of Technology Kharagpur, India}

\address[mymainaddress1]{Department of Computer Science, Mahishadal Raj College, West Bengal, India}

\address[mysecondaryaddress]{Cloud Computing and Distributed Systems (CLOUDS) Laboratory, School of Computing and Information Systems, \\ The University of Melbourne, Australia}

\begin{abstract}

Several researches and evidence show the increasing likelihood of pandemics (large-scale outbreaks of infectious disease) which has far reaching sequels in all aspects of human lives ranging from rapid mortality rates to economic and social disruption across the world. 
In the recent time, COVID-19 (Coronavirus Disease 2019) pandemic disrupted normal human lives, and motivated by the urgent need of combating COVID-19, researchers have put significant efforts in modelling and analysing the disease spread patterns for effective preventive measures (in addition to developing pharmaceutical solutions, like vaccine). In this regards, it is absolutely necessary to develop an analytics framework by extracting and incorporating the knowledge of heterogeneous data-sources to deliver insights in improving administrative policy and enhance the preparedness to combat the pandemic. Specifically, \textit{human mobility, travel history} and other \textit{transport statistics} have significant impacts on the spread of any infectious disease. In this direction, this paper proposes a spatio-temporal knowledge mining framework, named \textit{\textbf{STOPPAGE}} to model the impact of human mobility and other contextual information over large geographic area in different temporal scales. The framework has two major modules: (i) \textit{Spatio-temporal data and computing infrastructure} using fog/edge based architecture; and (ii) \textit{Spatio-temporal data analytics} module to efficiently extract knowledge from heterogeneous data sources. Typically, we develop a \textit{Pandemic-knowledge graph} to discover correlations among mobility information and disease spread, a deep learning architecture to predict the next hot-spot zones; and provide necessary support in home-health monitoring utilizing Femtolet and fog/edge based solutions. The experimental evaluations on real-life datasets related to COVID-19 in India illustrate the efficacy of the proposed methods.
\end{abstract}

\end{frontmatter}

\section{Introduction}
The significant growth of global travel, improved communication, urbanization, and greater exploitation of the natural environment have escalated the likelihood of outbreak of infectious diseases in a larger geographical scale. These fatal infectious diseases pose real threats for public health and government agencies require effective health measures in pandemic situations. In this context, the global outbreak of infectious disease COVID-19 (Coronavirus Disease 2019), caused by SARS-Cov-2, has swept 200+ countries or territories and contracted more than 160+ million people (as on the third week of March 2021). This human-to-human disease transmission is highly contagious and it is already observed that traditional infection-control or public-health measures to combat COVID-19 are inadequate. In the exigency situation, governments had already taken varied measures and policies such as travel-restriction, lockdown\footnote{Lockdown: International/ National border closures, restrictions of public activity
(school and business closures), and overall movement restriction whenever possible} of several regions, self-quarantine to control the rapid growth/ spread of the pandemic. However, still the threat of second wave of this pandemic persists in several countries of the world. 
\par The rapid development and emergence of Internet of Things (IoT) has significantly improved all life spheres and industries by connecting billions of devices and people around the world. IoT has brought useful solutions in different fields such as agriculture, farming, healthcare, smart building, smart city, personalized recommendation etc. \cite{reco}, \cite{xu2014ubiquitous}. The IoT devices need to send data to cloud servers frequently for processing and analysing the accumulated data. However, this increases the delay, therefore affects the Quality of Service (QoS). Here, edge and fog nodes extend the functionality of the cloud computing by processing, analysing and storing the information near the end-user. The combination of these emerging technologies facilitate several time-critical applications, namely, continuous patient monitoring in home, ambulance, and hospital, as well as assisting users in emergency situations, like disaster \cite{mahmud2020application}. 

\par In the recent times, there is a growing need of analysing spatio-temporal datasets for extracting meaningful information, and providing location-aware services, such as trip-planning, weather forecasting and even health-management.  Notably, one of the major aspects in epidemiological analysis is retrieving the correlations among the people and disease spread in spatial and temporal dimensions to effectively adapt the countermeasures. For instance, it was found in a spatio-temporal data analytics \cite{vinten2003cholera} that the source of Cholera outbreak (in London, 1854) was contaminated through bore wells. 
The finding was immensely helpful to combat the spread of Cholera. To this end, Internet of Spatial Things (IoST) combines IoT with spatial context \cite{eldrandaly2019internet}, where location information of the objects plays an important role. To fight against pandemic, spatio-temporal information and health data need to be integrated and analysed to predict the spread of the disease as well as assisting users about the risk of infected. Here, we propose the term \textit{Internet of Spatio-Health Things (IoSHT)} as an integration of IoST with IoHT (Internet of Health Things), where all the health-related information is combined with the location data. This work mainly focuses on how novel \textit{spatio-temporal data analytics method} is beneficial in deciding strategical administrative planning to enhance the preparedness to curb the pandemic. Undoubtedly, the entire world is suffering enormous amount of challenges from the pandemic situation caused by COVID-19:
\begin{itemize}
    \item (Q1) How to share information about available medical facilities and disease spread patterns in varied locations? 
    \item (Q2) What is the most effective measure to analyse available data sources and predict next probable hotspots to further enhance the preparation for pathological disease outbreak?
  \end{itemize}   
 
To resolve $Q1$, an efficient data infrastructure is required, which can store, manage and share authentic information amongst stakeholders. Further, novel spatio-temporal data analytics module is a must for understanding the impact of human movement data in the disease spread and summarizing the information in this situation ($Q2$). To this end, we propose a \textit{Pandemic-knowledge graph} (PKG) to capture the interdependence and connections among several entities and users' movement log, and subsequently identifying the hotspot/containment zones. The knowledge graph is represented as a multi-relational graph consisting of entities as nodes and relations as edges, and support several Artificial Intelligence (AI) related applications. However, it is difficult to represent mobility and other contextual information utilizing the conventional knowledge-graph that changes in temporal scale. On the other side, \textit{deep learning} is the most feasible solution to extract the correlations among several factors such as population density, mobility information and other data-sources.

\begin{table*}
 \centering
           \caption{Comparison of significant existing works and \textbf{STOPPAGE} framework in the context of COVID-19. \\ \textbf{A}: Efficient COVID-19 related knowledge retrieval and management; \textbf{B}: Movement data analysis to find out the impact; \textbf{C}: Deep learning module to analyse pandemic data; \textbf{D}: Cloud-fog-edge enabled framework deployment for faster response.}
    \resizebox{\textwidth}{!}{%
    \begin{tabular}{|c|c|c|c|c|c|c|} \hline
             SI & Publication & System & \multicolumn{4}{c|}{Feature}\\ \cline{4-7}
             & Reference & & A & B & C & D \\ \hline
             \multirow{5}{*}{[I]} & Wang et al. \cite{wang2020covid} & Comprehensive knowledge discovery framework to extract finegrained knowledge elements  & \ding{51} & \ding{56} & \ding{56} & \ding{56} \\ 
            & & from scientific literature, and a case study on \textit{Drug Repurposing
Report Generation} has been presented &  &  &  &  \\ 
            \cline{3-7} 
            & Esteva et al. \cite{esteva2020co} & Retriever-ranker semantic search engine to resolve complex queries to find scientific answers & \ding{51} & \ding{56} & \ding{56} & \ding{56} \\ \cline{3-7} 
            & Fernandez et al. \cite{domingo2020covid} & Cause-and-effect network generated from scientific literature and presented a knowledge graph. &  \ding{51} & \ding{56} & \ding{56} & \ding{56} \\
            & & The paper analyzes comorbidities, symptoms, and discovered over 300 candidate drugs for COVID-19 &  &  &  &  \\ 
            \cline{3-7} 
           & Chen et al. \cite{chen2020coronavirus} & Formalization and extraction of meaningful insights from the PubMed dataset and generation of &  \ding{51} & \ding{56} & \ding{56} & \ding{56} \\ 
           & & knowledge graph using two methods: co-occurrence frequency and cosine similarity  & &  &  &  \\ \cline{3-7} 
           & Reese et al. \cite{reese2020kg} &  Present flexible framework to integrate biomedical data for producing knowledge graphs (KGs)  &   \ding{51} & \ding{56} & \ding{56} & \ding{56} \\ \hline
            \multicolumn{7}{|l|}{\textbf{\textit{   [1] Most of the works deal with knowledge mining related to pharmaceutical data, drug discovery or extracting information from scientific literature.}}} \\ 
            \multicolumn{7}{|l|}{\textbf{\textit{\hspace{0.3cm}No existing works deal with information storage and knowledge extraction from heterogeneous data sources, like travel-logs, aggregate movement }}} \\ 
             \multicolumn{7}{|l|}{\textbf{\textit{\hspace{0.3cm}information, COVID-19 statistics in different spatio-temporal granularity and find the correlations amongst them.}}} \\ 
            \hline \hline 
           \multirow{5}{*}{[II]} &  Kraemer et al. \cite{kraemer2020effect} & The spatial distribution of COVID-19 cases in China and the mobility traces are analyzed & \ding{56} & \ding{51} & \ding{56} & \ding{56} \\ \cline{3-7} 
           & Pepe et al. \cite{pepe2020covid} &  The daily time-series of three different aggregated mobility metrics have been presented. The   & \ding{56} & \ding{51} & \ding{56} & \ding{56} \\ 
            & & authors use 170,000 smartphone users'  data to monitor the impact and assisting in making decisions. &  &  &  &  \\ \cline{3-7}
          & Huajun et al. \cite{he2020efficient} & Identification of suspected infected crowds based on the human movement trajectories. & \ding{56} & \ding{51} & \ding{56} & \ding{56} \\ \cline{3-7}
         & Samuel et al. \cite{engle2020staying} & Analyse human mobility data to find out disease spread for restrictions order to stay home. & \ding{56} & \ding{51} & \ding{56} & \ding{56} \\ \cline{3-7}
          & Hamda et al. \cite{badr2020association} & A mathematical modelling with the movement dynamics and disease outbreaks have been proposed. & \ding{56} & \ding{51} & \ding{56} & \ding{56} \\ 
          & & The work helps in taking measures of social distancing and other policies in the 25 counties in the USA &  &  &  &  \\ \hline  \hline 
         \multirow{3}{*}{[III]} & Rahman et al. \cite{rahman2020b5g} &  Distributed deep learning framework for COVID-19 diagnosis in 5G network at the edge & \ding{56} & \ding{56} & \ding{51} & \ding{56} \\           \cline{3-7}
          & Kapoor et al. \cite{kapoor2020examining} & Forecasting approach using Graph Neural Networks and mobility data for COVID-19 case prediction. & \ding{56} & \ding{56} & \ding{51} & \ding{56} \\             \cline{3-7}
        & Luz et al. \cite{luz2020towards} & An efficient method of COVID-19 screening in chest X-rays in less time and computational cost. & \ding{56} & \ding{56} & \ding{51} & \ding{56} \\         \hline
          \multicolumn{7}{|l|}{\textbf{\textit{\hspace{0.3cm} [II]-[III] To the best of our knowledge, there is no deep learning module which analyzes the movement patterns and other contextual facts to understand}}} \\
           \multicolumn{7}{|l|}{\textbf{\textit{\hspace{0.3cm} the impact of the disease spread and predicts hotspots efficiently.}}}  \\\hline 
           \multirow{4}{*}{[IV]} & Tuli et al. \cite{tuli2020predicting} & Cloud computing based framework to predict growth of the epidemic and appropriate & \ding{56} & \ding{56} & \ding{56} & \ding{51} \\   
             & & strategies have been presented. The authors present a prediction framework. & & & & \\ \cline{3-7}
             & Wang et al. \cite{wang2018anonymous} & Anonymous and secure aggregation model in fog-based public cloud computing model & \ding{56} & \ding{56} & \ding{56} & \ding{51} \\   \cline{3-7}
             & Md et al. \cite{whaiduzzaman2020privacy} & Utilize mobile and fog computing to trace and prevent COVID-19 community transmission   & \ding{56} & \ding{56} & \ding{56} & \ding{51}   \\ \cline{3-7}
             & Adarsh et al. \cite{kumar2020drone} & Multi-layered architecture to collect real-time information from drones and utilize in & \ding{56} & \ding{56} & \ding{56} & \ding{51} \\  
             &  & disease monitoring, control, thermal imaging, social distancing, and statistics generation & & & & \\ \hline \hline 
            &  \textbf{STOPPAGE} & Cloud-fog-edge enabled collaborative framework conducive to  construct a novel time-dependent & \ding{51} & \ding{51} & \ding{51} & \ding{51} \\  
            & (Proposed framework) & Pandemic-Knowledge graph (PKG) based
on the spatio-temporal data. Presenting deep learning based  &  &  &  &  \\ 
& & analytics module to find out hotspots by mining contextual data.  &  &  &  &  \\ \hline
                        \end{tabular}%
            }
\label{relwork}   
\end{table*}

\par Although, the full lockdown measure (complete restriction of the movements) is one of the optimal solution to reduce the human-to-human transmission rate, on the counter side, it has a drastic impact on the economy of the country. This paper analyses the impact of the movement in two phases in the context of India: (a) In the pre-lock down phase when the International travel was allowed. It analyses the in-flow of the International flights in several regions of the country, and the cases reported within a temporal buffer of the visits. The approximate count of the travellers and air-crafts in the time span of March $10$ to March $21$, 2020\footnote{The international flights were suspended in India from March 22, 2020 onwards.} and the reported cases\footnote{\url{https://api.covid19india.org/}} have been analysed; (b) In the next phase, we develop a disease spread module based on the factors such as population density, changes of mobility patterns\footnote{Google mobility data: \url{https://www.google.com/covid19/mobility/}}, POI (point-of-interest) information etc. Here, we have proposed a variant of co-occurrence pattern where the mobility information has been augmented. 
\par The major significance and contributions of \textbf{STOPPAGE} are as follows:
\begin{itemize}
 \item \textbf{IoSHT}: An end-to-end spatial data infrastructure framework named, IoSHT, consisting cloud, fog, edge and IoT layers is presented which stores, manages and analyses pandemic related data effectively and assists in taking effective decisions. IoSHT is capable of collecting BAN (Body area network) information and other contextual data such as environment-temperature, location data, mobility information etc. Femtolet (a small cell base station with storage and computation ability) \cite{mukherjee2016femtolet} is used for faster reporting of patient information of hospitals. The delay and power consumption of the user device for health status monitoring have been determined and compared with the cloud only system to show the efficacy of the proposed framework.
 
    \item \textbf{Pandemic-Knowledge Graph (PKG):} We propose a novel time-dependent {\em Pandemic-knowledge graph} (\textit{PKG}) based on the active cases, POI, road and air-travel connectivity index and users' movement history. \textit{PKG} captures correlations and impacts among locations, temporal information, movement semantics and infectious disease spread.
    
    \item The paper proposes a \textbf{deep learning network} which incorporates the spatio-temporal data instances and finds out the probable next hotspot zones by mining the inherent knowledge. To be specific, the deep learning architecture is capable to incorporate the impacts of human mobility and other spatio-temporal features of PKG, and finds out the hotspots effectively.

    \item The experimentation have been carried out using the real-life datasets available in India. The road-network, POI information and air-traffic datasets have been collected from OpenStreet Map (OSM), Google Map Services and crawling web-pages of Airports Authority of India (AAI). The information about the testing facilities and hospitals have been collected from Indian Council of Medical Research\footnote{ICMR:\url{https://www.icmr.gov.in/}}. The changes of mobility information have been analysed from Google Mobility Report. Finally, we have also accumulated individuals' mobility history from Google Map Timeline from specific regions of India to evaluate the efficacy of \textit{STOPPAGE} framework. 
\end{itemize}
The rest of the paper is organized as follows. Section 2 summarizes the related works in this domain. Then, the system architecture and the deep learning module have been presented in section 3. Then we present a real-life example of home-health monitoring system and usages of Femtolet to reduce the reporting time in section 4 followed by the experimental evaluations in section 5. Finally, we conclude the paper with research challenges and opportunities in this domain in section 6.

\section{Background and Related Works}
To begin with, Internet of Spatio-Health Things is an emerging field of IoT, where integration of IoHT and IoST is carried out to improve the health care system. In IoHT the IoT devices are integrated with mobile technologies to process and exchange data for monitoring health condition of individuals \cite{da2018internet, santos2020online}. 
While several IoT-enabled medical devices, BAN accumulate health data and provide insights into symptoms and trends, the information-exchange, communication and interoperability of IoT devices make healthcare services more effective. Inside the health-care centers, IoHT can enable vital signs monitoring of a patient \cite{da2018internet}. However, the COVID-19 pandemic situation can not be managed by only IoHT solutions, it also requires the monitoring of human-to-human transmission of the disease. To this end, IoHT needs to be integrated with IoST to monitor the overall spread of the disease and reduce the transmission rate by early identification and taking preventive measures, such as restricting human movement or close contact. Further, IoT devices associated with real-time location tracking  of medical equipment such as oxygen pumps, wheelchairs, defibrillators make medical resource management efficient in this pandemic. It is evident that location-information has a huge significance to curb the pandemic. Specifically, \textit{mobility traces} play a pivotal role in several real-life applications, such as, location-prediction \cite{jnca2,jncanow2}, outdoor navigation of visually impaired people \cite{jncanow1} etc. It is quite obvious mobility is also an important factor for spreading infectious diseases. \textit{Internet of Spatial Things (IoST)} is a new domain and only a very few works (e.g. \cite{eldrandaly2019internet, ghosh2019mobi}) have been carried out in this direction. In this paper, we introduce \textit{Internet of Spatio-Health Things (IoSHT)}, where individual health monitoring and assistance can be provided anytime anywhere, as well as the information-sharing, pandemic/disease spread pattern are analysed to enhance the preparedness.
\begin{figure*}
    \centering
    \includegraphics[scale=0.80]{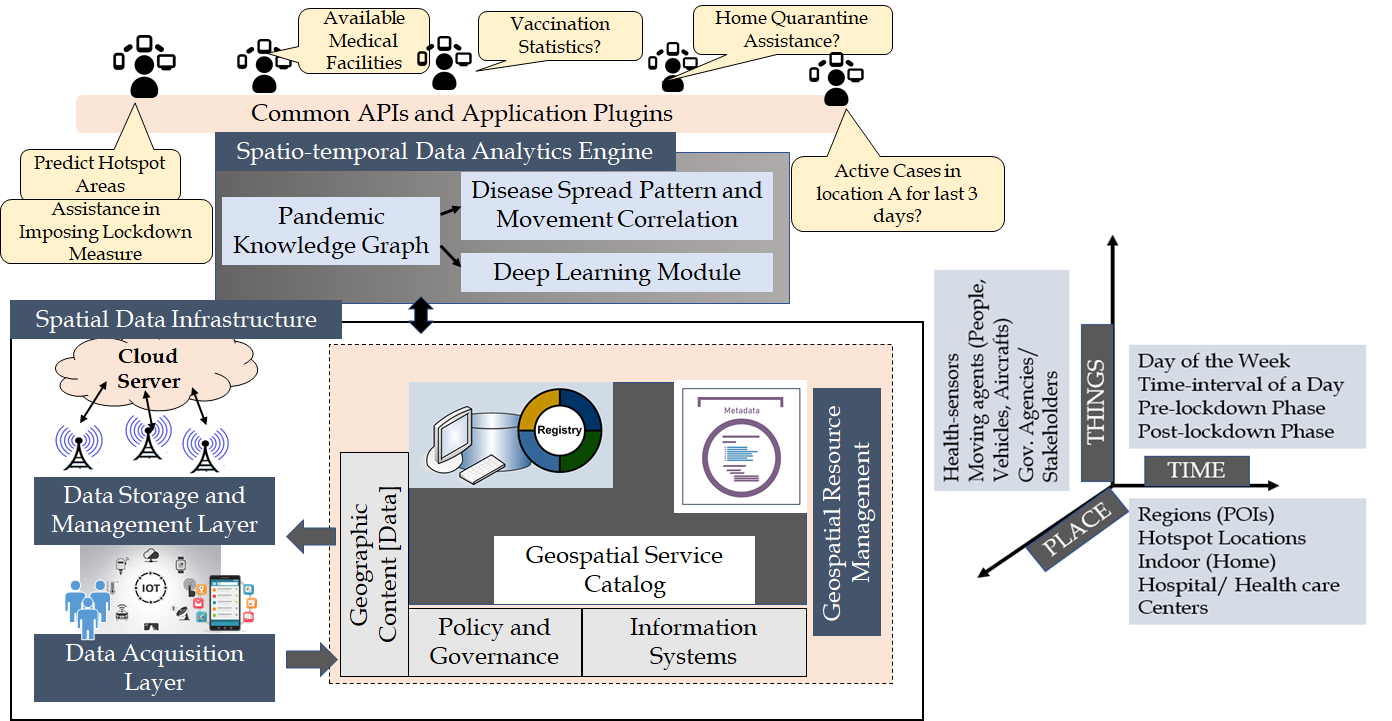}
    \caption{\textbf{STOPPAGE}: Overall Workflow}
    \label{overallstop}
\end{figure*}
\par The onset of COVID-19 has resulted large volume of research papers proposing several methods to curb the pandemic. While there are numerous articles dealing with the pharmaceutical methods, we emphasize more on analysis of contextual information collected from mobile-phones and other sources, such as movement patterns, cumulative count etc. to predict the future spread of the disease. Table \ref{relwork} shows four major features and the existing approaches in the context of COVID-19. Data management and information retrieval is one of the major step in combating the pandemic. A comprehensive insight on how IoMT (Internet of Medical Things) systems can be beneficial in the context of COVID-19 along with the architecture, tools and technologies are discussed in \cite{aman2020iomt}. Marcello et al. \cite{ienca2020responsible} states that large scale collection of data and maintaining the privacy and public trust is one of prime step. There are several works on summarizing the knowledge from 
drug discovery or other pharmaceutical methods \cite{domingo2020covid}, \cite{chen2020coronavirus}. Also, researchers \cite{wang2020covid} have put significant efforts to summarize information from scientific literature by proposing knowledge graph. There are works on aggregate movement dynamics and understanding the impact in disease spread \cite{kraemer2020effect}, \cite{pepe2020covid}. Huajun et al. \cite{he2020efficient} present a novel trajectory pruning method to find out suspected crowds in a region. However, to the best of our knowledge, no existing works has extensively studied the movement patterns and contextual information in a region in the context of COVID-19 and effectively stores and manages information for better decision making. There are also few works leveraging cloud-fog paradigm for contact tracing, real-time drone based system and analysing the growth of the disease \cite{tuli2020predicting}, \cite{whaiduzzaman2020privacy}. 
\par It may be concluded that machine learning and movement data analytics may act as  major scientific tools for combating the spread of COVID-19 pandemic. In this direction, our work explores and analyses the human movement related data proposing novel knowledge graph (PKG) and deep learning architecture.

\section{STOPPAGE: System Architecture}

\begin{figure}
    \centering
    \includegraphics[scale=0.5]{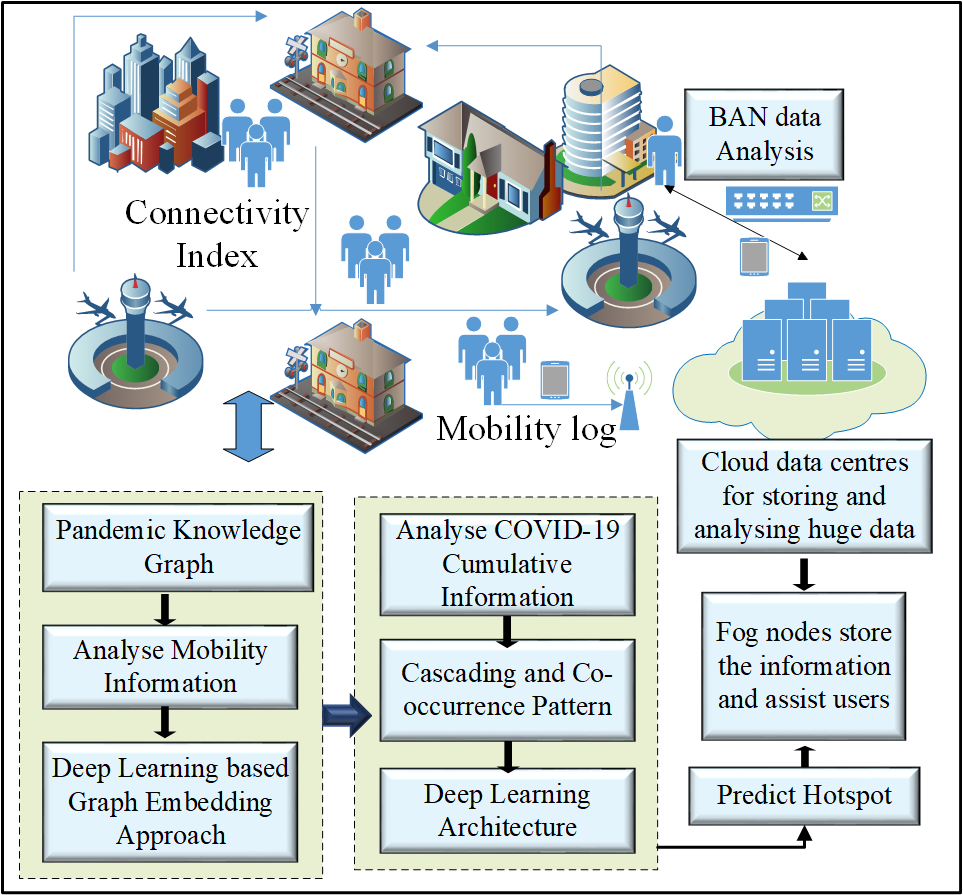}
    \caption{\textbf{STOPPAGE}: Spatio-temporal Data Analytics Module}
    \label{overall}
\end{figure}

The proposed framework, \textit{STOPPAGE} has two major modules, namely: spatio-temporal data infrastructure (SDI) and a data analytics engine running over the SDI. Figure \ref{overallstop} illustrates the overall workflow and modules of \textit{STOPPAGE}. The spatial data infrastructure consists of backbone IoT network and geospatial resource management module. The spatio-temporal data analytics engine analyses the COVID-19 data and mobility related information to find out the probable hotspot zones by deploying deep learning architecture. Furthermore, \textit{STOPPAGE}proposes the use of Femtolet to report any health events in minimal latency. Also, the health status monitoring is provisioned in the framework by a common API-endpoint. Fig. \ref{overall} illustrates the overall block-diagram of data analytics engine of STOPPAGE. It has three major components (i) Pandemic knowledge graph (PKG) construction, (ii) deep learning based movement analytics and (iii) faster reporting of COVID-19 events and data sharing using cloud-fog-edge based paradigm. Fig. 3(a) represents the SDI (Spatial Data Infrastructure) and Fig. 3(b) illustrates the deployment of IoSHT in cloud-fog-edge structure. 
\begin{figure*}[ht]
    \centering
    \includegraphics[scale=0.64]{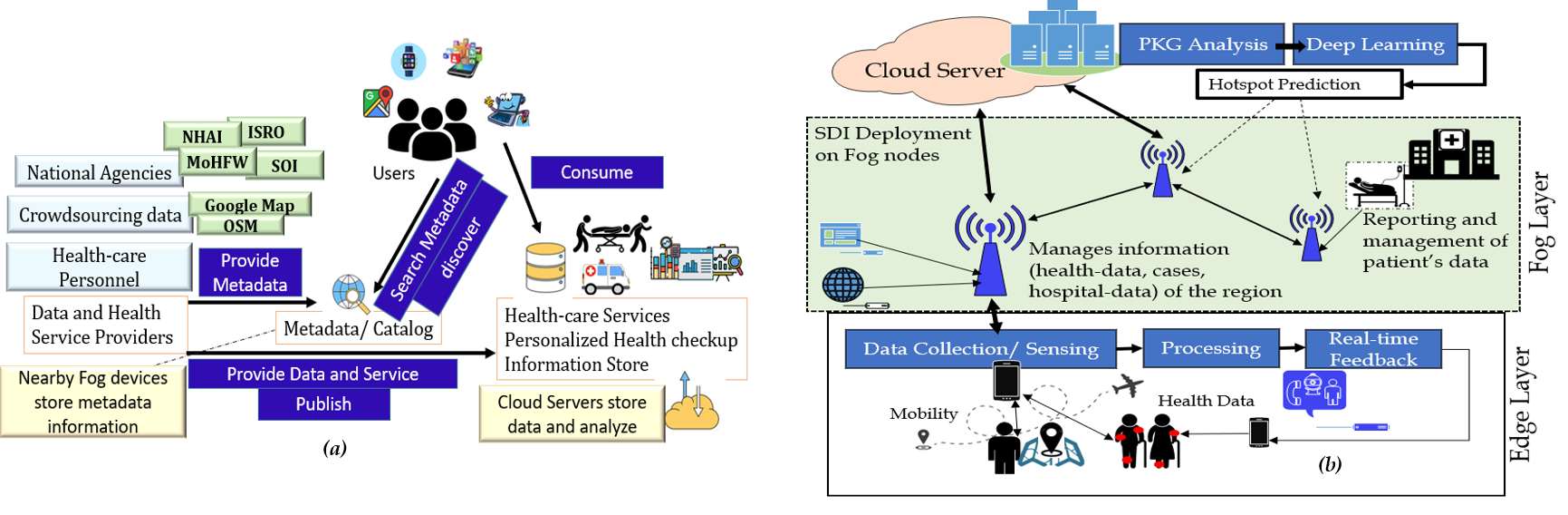}
    \caption{(a) SDI (Spatial Data Infrastructure) in the context of IoSHT, (b) Cloud-Fog-Edge hierarchical structure}
    \label{sdi}
\end{figure*}
\subsection{Spatial Data Infrastructure (SDI): An Enabler of IoSHT}
 \textit{Spatial data infrastructure (SDI)}\footnote{The term Spatial Data Infrastructure (SDI) was coined in 1993 by the U.S. National Research Council} denotes a framework consisting technologies, policies and orchestration method to create, exchange, and utilize geospatial information and services among the community \cite{goodchild2007citizens}. The goals of this framework are:
 \begin{itemize}
     \item ease of search and discovery of geospatial services and information 
     \item reduce data duplication of information among the national agencies (government) 
     \item seamless data sharing technique
     \item maintaining the data integrity and privacy
     \end{itemize}
Further, SDI enables the communication between the repositories of cities, states, countries and industries to share the data and services. In other words, the difficulty in sharing and accessing different format/structure of data is resolved by SDI. In the context of COVID-19, different stakeholders, such as health-department, transportation authority or epidemic-control team along with the end-users need to seamlessly share data and communicate to effectively handle the situation. Without a proper underlying infrastructure of data sharing and managing mechanism, any epidemic-control measure or policy is difficult to implement. Therefore, in our proposed IoSHT framework, we have adapted SDI as the backbone infrastructure. The SDI in the context of IoSHT is represented in Fig. \ref{sdi}(a). To combat the disease spread and take countermeasures, we need to analyse heterogeneous data-sources. The authentic data about the country's population, demography, health care centers etc. can be found from the government departments. In the context of India, we collect demography and population data from SOI and open-source platform\footnote{Survey of India: \url{http://www.surveyofindia.gov.in/}},\footnote{\url{https://www.diva-gis.org/gdata}}, health-related information from MoHFW\footnote{Ministry of Health and Family Welfare of India: \url{https://www.mohfw.gov.in/}}, transport data from NHAI\footnote{National Highways Authority of India: \url{https://nhai.gov.in/}} and raster information (satellite data) from ISRO\footnote{Indian Space Research Organisation: \url{https://www.isro.gov.in/}}. All of these datasets of country's population, demography information of the people, health care center related information, traffic data are useful to analyse the disease spread patterns. Further, the crowdsourcing data from Google Map, OpenStreetMap (OSM), or Android application are also beneficial to get the present status of the region. The data and health service providers provide the metadata information to the catalog. This catalog is maintained in each of the nearest fog devices, where the data or the service is generated. User can search the catalog any time, and extract the metadata of the information or service he/she requires. Once, the user selects the required service/information, he/she can consume the service. It may be noted, in the proposed \textit{STOPPAGE} framework, the services are: (i) contact the health care center for medical help; (ii) personalized health-checkup by sending the health-parameters' values collected from the IoT devices/BAN; (iii) get the present disease spread information of any region; (iv) find the risk of infection analysing his/her movement history. In our model, the metadata information is retrieved in \textit{.xml} or \textit{.json} format. All of these analytics are computed in the cloud servers, and the user gets the result through the web-dashboard or android app.   
\subsection{Health status Monitoring and Reporting using Fog-Edge Solution}
In the previous section, we have discussed the proposed system architecture of \textit{STOPPAGE}. For current health status prediction of people in a region, we have to consider the reports of health care centres/hospitals. Here, Femtolet is used in the health care centers for faster reporting of the number patients affected, deceased and cured, because this information is also very important to monitor the current status of the pandemic in a geographic region. Femtolet is small cell base station which has storage and processing ability \cite{mukherjee2016femtolet}. 
 \begin{figure}
    \centering
    \includegraphics[scale=0.45]{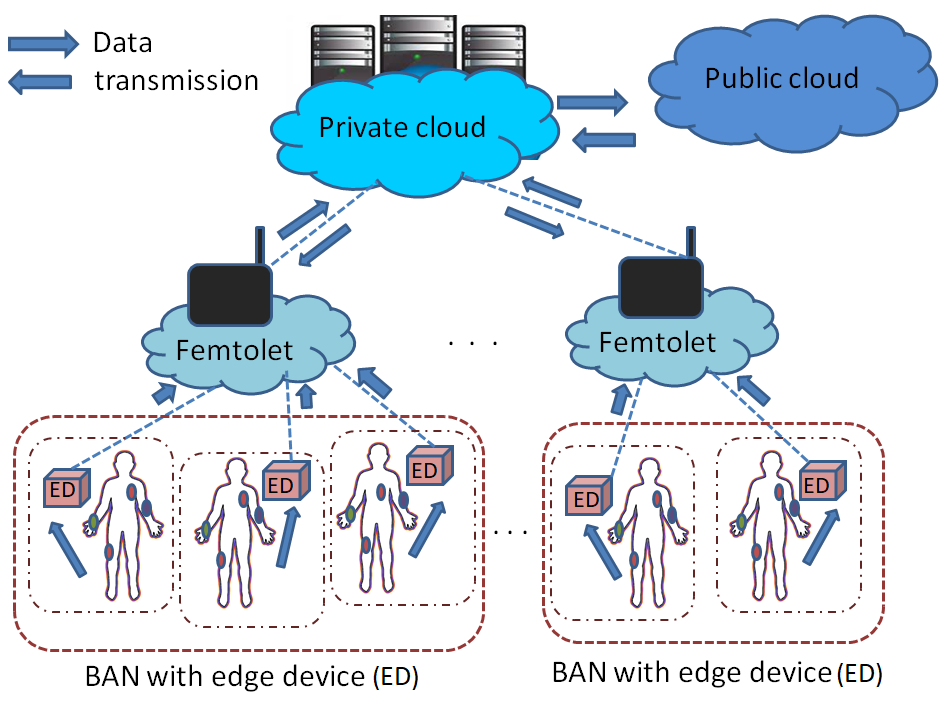}
    \caption{Smart health care framework based on Femtolet}
    \label{fig:femtolet}
\end{figure}

In the proposed model, in each ward of the hospital Femtolets are deployed to store and process the information of the patients currently admitted in that ward. The number of Femtolets to be allocated depends on the coverage area, storage capacity, data processing speed of the Femtolet, and the number of patients currently admitted in that ward. The health status of a patient is collected using BAN and sent to an edge device (for each patient BAN and an edge device are used). The edge device (ED) accumulates the data and sends to the Femtolet, connected to the edge device. The Femtolet processes the incoming data and stores it under the respective patient registry (for each patient a data registry is maintained). If any emergency situation arises, the Femtolet sends signal to the connected alarm which after receiving the signal starts ringing. The health personnel of that ward then takes required action. The Femtolet periodically sends the patient health status to the private cloud servers of the health centre, which maintains the information regarding the number of patients cured, dead and affected. From the private cloud servers the details of the number of patients affected, cured, and deceased in the hospital with respect to the disease are reported to the public cloud servers periodically. The overall architecture of Femtolet based health care model is shown in Fig. \ref{fig:femtolet}. As the public cloud has the information regarding the number of patients admitted, cured, and dead due to COVID-19 in each hospital, the total number of admitted, cured and dead patients (considering all the hospitals' information) due to COVID-19 is determined. The people can get this information if he/she wishes to know the same. However, the information of each hospital will not be disclosed without authorization due to privacy management. The cloud already has the information regarding hotspot zone (in Section 3.4 hotspot identification is discussed). By considering both the information, the people can be made aware of the current scenario.

\subsection{Spatio-temporal Data Analytics Engine: Pandemic-Knowledge Graph (PKG) Construction}
This section describes the features of \textit{Pandemic-Knowledge graph (PKG)} and how it is constructed from the available dataset. This is a multi-layer network - which captures the movement behaviour individuals as well as the overall movement semantics of the ROI (Region-of-interest) and the statistics related to COVID-19, such as growth pattern of active cases, changes of aggregated mobility patterns, available medical facilities etc. The intuitions behind constructing knowledge graph are: 
   (i) Transforming the underlying relations of mobility semantics in a machine-readable format to support information retrieval and query-processing,
    (ii) The complex spatio-temporal mobility dataset can be represented by graph structure effectively instead of other storage, 
    (iii) Updation of relations and facts can be easily incorporated in this structure compared to SQL-based processing, therefore provides more flexible schema, and finally 
    (iv) Knowledge graph provides a higher level abstraction of information, which may help to extract more complex and previously unknown interrelations merging more than one relations of the graph. Therefore, it facilitates building a semantically enriched knowledge-base.   

\vspace{-0.1in}
\subsection*{Pandemic-Knowledge Graph (PKG)}
The Pandemic-Knowledge graph is formulated as triplet of $<s,r,o>$, where $s \subseteq \Omega$ and $o \subseteq \Omega$ are \textit{entities} and $r \subseteq \varphi$ is the relation between two entities. The sets of entities and relations are denoted by $\Omega$ ($\Omega_1 \cup \Omega_2$) and $\varphi$ respectively. Notably, in our Pandemic-Knowledge graph proposition, each triplet (or fact) has a time-slot, when the fact is valid. Therefore the facts of $PKG$ takes the form of $MF:<s,r,o,[t_1,t_2],f>$ where the $[t_1,t_2]$ entry denotes the time-interval when the fact is true. $f$ is the feature value of the relation. This $f$ is introduced to capture the semantic correlation with the COVID data.  

    The entities consist of: $U \subseteq \Omega_1$ and $P \subseteq \Omega_2$, where $U$ and $P$ are the set of users and set of place-ids of a ROI. 
    Each entity $pl \subseteq P$ has attributes: POI-type, location (latitude and longitude), enclosing area. Few $pl$ has additional features such as \textit{opening/closing time}, available facilities. 
    Each entity $u \subseteq U$ has unique user-id, age, gender, residence-area, health-profile and travel-history. 
    The derived attributes of POIs such as \textit{GPS footprint} or \textit{edge segment} are extracted from the GPS log of the users in the ROI. The facts \textit{connectedBy} and \textit{boundingBox} can be directly computed from the road-network and latitude and longitude information of the POI.  
    Based on the historical movement log the relations $r \subseteq \varphi$ are extracted. 
Apart from the mobility fact label, each edge in the graph has a feature value associated, which depicts the probability of the occurrence of the edge. Few such facts are represented as follows:

\vspace{-0.2in}
\begin{equation}
    visit: MF(u_a,visit, p_i,[t_1,t_2],f_x)=TRUE
\end{equation}
User $u_a$ visits place $p_i$ in the time-interval $[t_1,t_2]$ with a probability of being infected $f_x$. Here, the feature value of the fact is computed by the graph embedding approach \cite{baselinekg}. \vspace{-0.1in}
\begin{equation}
\begin{split}
    group: MF(u_a,group, U' \setminus u_a ,[t_1,t_2],f_x)=TRUE \\
    \exists (u_1,\dots,u_n) \exists (p_1,\dots,p_m) \forall_j ^n \\
    [MF(u_j,visit, p_1,[t_1,t_1^{'}],f_1) \wedge MF(u_j,visit, p_2,[t_2,t_2^{'}],f_2) \\ 
    \wedge \dots \wedge MF(u_j,visit, p_m,[t_m,t_m^{'}],f_m)]=TRUE,  \; \; m \geq 3 \\
\end{split}
   \end{equation}
  The sequence of places $(p_1, \dots, p_m)$ are visited by a group of users $(U')$ in the same time-intervals. $f_x$ may have a value within [0,1). 
  
  \begin{equation}
\begin{split}
   flow: MF(P_1, flow,  P_2 ,[t_1,t_2],f_x)=TRUE \\
    \exists (u_1,\dots,u_n) \forall_j ^n [MF(u_j,visit, p_a,[t_1,t_1^{'}],f_x) \\ \wedge MF(u_j,visit, p_b,[t_2,t_2^{'}],f_x)=TRUE, \; \; n \geq \nu \\
\end{split}
   \end{equation}
  The above mentioned movement flow (MF) is detected from $p_a$ to $p_b$ in a specific time-slot, when a set of users ($u_1, \dots, u_n$) visit the POI $p_a$ followed by $p_b$ in the same time-interval. The count ($n$) of GPS footprints satisfying the fact must be greater than a threshold value $\nu$. In other words, if $\geq \nu$ people visit the sequence of POIs in same time-interval, then it is considered that there is a mobility flow from $p_a$ to $p_b$. 
  \begin{equation}
  \resizebox{.5\textwidth}{!} 
{      
$ hotspot: MF(boundingbox(\{p\}), infected, \{u\}, [t_1, -],co)=TRUE $
 } 
 \end{equation}
  The bounding box of the set of regions (${p}$) is a hotspot with $co$ number of infected people. 
\begin{equation}
    connectivity: MF(p_a,connect, p_b,[t_1,t_2],f_p)=TRUE
\end{equation}
The place $p_b$ can be visited from place $p_a$ in the time-interval $[t_1,t_2]$ where $f_p$ denotes the number of available routes. It may be noted that due to the lockdown measure, this fact changes frequently for all pairs of places. The connectivity index ($CI$) of a place $p$ is computed based on the \textit{connectivity} and the in-flow and out-flow of traffic from a particular place $p$. The high connectivity index between two place represents higher mobility flow, and thus can be a potent feature for the hotspot detection. 
Next, we define \textit{cascading pattern} ($P_{ca}$) and \textit{co-occurrence pattern} ($P_{co}$) extracted from PKG. These patterns can be analysed from the travel history of the persons who are detected COVID-19 positive. Cascading patterns represent events whose instances are located together and occur serially. For instance, analysing the PKG data of a hotspot may reveal that a \textit{group} of users \textit{visit} a POI on a day and spends a particular time-duration within a distance threshold. Also, air-travel of a group of users from a region with high number of active cases and new hotspot emergence in the destination region within a time-span is another example of cascading pattern. Also, $P_{ca}$ covers patterns such as: higher number of medical facilities in a city-region, more cases of COVID-19 patient transfer and emergence of hotspot.   
On the other side, \textit{co-occurrence patterns} reflects patterns occurring frequently within a spatial
range and a temporal span. This patterns comprise properties from different contexts. For example, given three different contexts, namely, population density, aggregate movement pattern and COVID-19 active/ new cases (say, within a spatial buffer of 2km and time-span of 7days):
$P_{co}:\{population-density > \delta; aggregate\; movement\; pattern > \gamma; Hotspot\_prob=HIGH\}$. Also, $P_{co}$ covers patterns such as: festival/ gathering events in a region and emergence of new hotspot; or more amount of location-based service requests (ridesharing services/ food delivery services) and hotspot detection in the same region within a temporal bound. Algorithm 1 shows the steps of extracting cascading and co-occurrence patterns from PKG, which in turn helps in extracting the knowledge of disease spread, and finding the possible hotspot zones in future. The algorithm needs a threshold value called \textit{pattern participation index ($PI$)} to prune the irrelevant patterns.
\begin{equation}
\begin{split}
    PI=\min [Pr(P_{ca}|MF) \; or \; Pr(P_{co}|MF)]  \\
    = \min [\frac{facts \; or \; events \; participating \; in \;P_{ca} \;or \;P_{co}}{Total \;no. \;of \;facts \; or \; events \; in \;PKG} ]
\end{split}
    \end{equation}
    Here, $MF$ represents the mobility facts and events present in PKG. It may be noted that the conditional probability of a pattern; given the possible similar type of events/ facts is measured by $PI$. It is also an useful measure for predicting the near future occurrence of a pattern in the spatio-temporal proximity of an observed instance of a participating event-type. We set two threshold values namely $PI_1$ and $PI_2$ for pruning the irrelevant patterns from PKG. The neighbor spatio-temporal or ST relation represents the time and spatial constraint on finding the co-occurrence patterns. In brief, the algorithm first extracts the plausible candidates from PKG, followed by pruning technique to eliminate irrelevant patterns. 
 \begin{algorithm}[!htb]
\caption{Extracting Cascading and Co-occurrence patterns from PKG}
\label{algo1}
\begin{flushleft}
 \textbf{Input:} Pandemic Knowledge Graph ($PKG$), neighbor ST relation ($NR[]$), set of event/ fact types ($MF$), pattern participation index thresholds ($PI_1, PI_2$) \\
 \textbf{Output:} Set of $P_{co}$ and $P_{ca}$ patterns
 \end{flushleft}
\small{
 \begin{algorithmic}[1]
 		\State $P_{ca}[\;] \leftarrow NULL\; and \;P_{co}[\;] \leftarrow NULL$  \Comment{\texttt{Create NULL arrays for storing patterns}}
		        \For{$i = 1$ to $size$} \Comment{\texttt{size: maximum length of the patterns}}
           \State{filter$(P_{ca}[i], PI_1)$} \Comment{\texttt{Remove candidates by analysing threshold value}}
           \State{Genpat(i) $\leftarrow P_{ca}\; of \; length\;(i-1)$} \Comment{\texttt{Extract new patterns of i length from patterns with length (i-1)}}
           \State $temp \leftarrow ST-Join(P_{ca}[i], NR)$ \Comment{\texttt{Perform spatio-temporal join with instances from direct neighbor relation}}
           \If{$checkThreshold(temp)\geq PI_1$}
           \State{$P_{ca}[i] \leftarrow Append(temp)$}  \Comment{\texttt{Append the new pattern if it satisfy the threshold condition}}
           \EndIf
                  \EndFor
            \For{$t=1 \; to (n-1)$}
            \State{$te \leftarrow GenCand(PKG(MF,NR))$} \Comment{\texttt{Generate candidates from PKG using possible facts/ events and neighbor relation}}
            \State{$P_{co}[t] \leftarrow sweepLine(te,P_2)$}\Comment{\texttt{Sweep-line based method to extract events satisfying ST-property and threshold}}
               \EndFor   
        
\end{algorithmic}}
\end{algorithm} 
\par Construction of the Pandemic-Knowledge graph consists of following steps:
(i)	In the first step, the POIs and users along with the attributes are extracted and stored. The information are discovered by geo-tagging step and segmenting the trajectory of individuals at different time-scales. 
(ii)	Next, the links between the entities (or facts) at different time-instances are discovered by knowledge graph embedding approach. The relations such as visit, group, flow etc. are defined. The movement log is analysed to measure the plausibility of any such facts.  However, the facts are checked over different temporal instances, since the facts or relations of the knowledge graph change with time-instances. The backbone SDI helps in sharing heterogeneous data, maintaining the authenticity and integrity of the data to build PKG, and facilitating services seamlessly to the user.  
\subsection{Spatio-temporal Data Analytics Engine: Deep Learning Architecture to find out Hotspots}
\textit{``You cannot fight a fire blindfolded. And we cannot stop this pandemic if we don’t know who is infected."}\footnote{WHO Director-General's opening remarks at the media briefing on COVID-19, dated March 16, 2020} Until there is no vaccine of this contagious disease to curb the pandemic, identifying the affected person and restricting the mobility to reduce human-to-human transmission is the only optimistic solution. 
Analyzing spatio-temporal pattern of disease outbreak is an integral part of pandemic control, which helps in identifying spatio-temporal hotspots (disease emerging areas), and subsequently assists the
planning of emergency measures for monitoring, surveillance, and prevention of the
disease spread. Most of the governments have enforced partial or complete lockdown to prevent the disease spread. However, the complete lockdown measure has negative impact on the socio-economic condition of a country, and more feasible solution is required, such as identifying high-risk zones (large number of affected person) and impose lockdown at those regions. In such a scenario, it is absolutely necessary to find the correlation of the disease spread and other additional information from spatio-temporal context. For instance, whether the disease spread is related to population density, demography data etc. Moreover, whether specific regions like, railway junction, commercial area is more susceptible to infection - needs to be analyzed. 

\subsection*{Predict the hotspots areas}
The primary objective of our analytics module is to identify the high-risk areas of disease spread in a region. We aim to find out the correlation of contextual information such as mobility with the probable spread of the disease. 
\par Of late, \textit{deep learning} has gained significant research interest to utilize the correlations between related tasks and improve the classification accuracy by jointly learning more than one task. Inspired by this paradigm, \textit{STOPPAGE} models the deep learning module to predict the hot-spot areas. It may be noted, that in our problem set-up, the mobility patterns have significant impact, and learning the representation of these contextual variables is one of the important aspects of our deep learning architecture. 
Fig. \ref{fig:deepl} illustrates the recurrent architecture of the proposed deep learning module.
 In our framework, we have considered the following features:
 \begin{figure}
    \centering
    \includegraphics[scale=0.65]{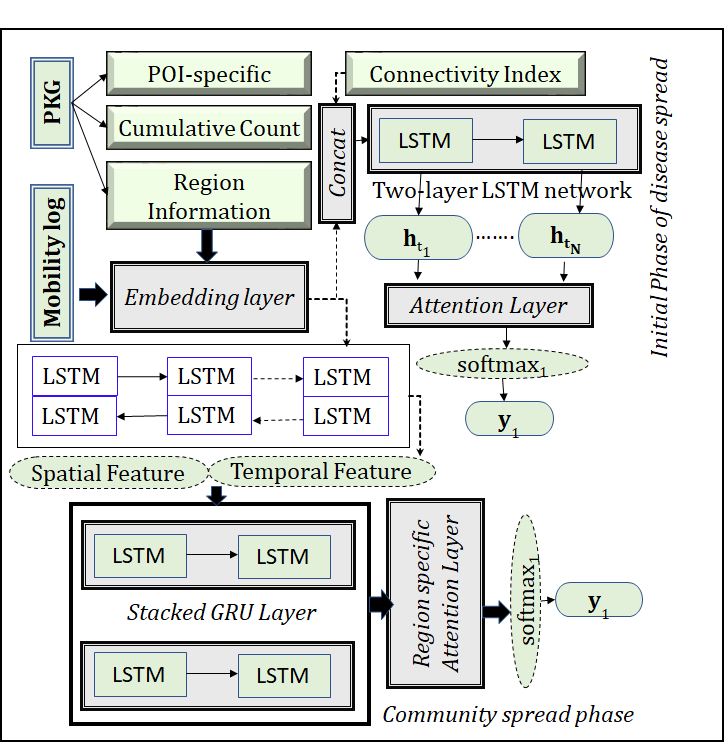}
    \caption{Deep Learning Architecture of \textbf{STOPPAGE}}
    \label{fig:deepl}
\end{figure}
\begin{enumerate}
    \item Demographic data ($D$): Demographic data is the statistical data collected about the characteristics of the population. We have considered $age$, $gender$, $employment\; status$ and $literacy\; rate$ of people of a region. 
    \item Population density ($PD$): It provides the number of people in each region along with the condition of their household (slum area/houseless etc.). The datasets $D$ and $PD$ have been collected from the census information board of India\footnote{\url{https://www.censusindia.gov.in/(S(esr3lm45pksguc451d45sp55))/2011census/population_enumeration.aspx}}, where district and zone-wise (specific locality) population and demographic data are available.
    \item Point of Interest (POI): The POI information, such as the locations of the airports, railway junctions, hospitals, and other commercial areas are analysed to extract insights into the possible disease spread pattern and in identifying the hotspot/containment zones. The data has been extracted from OSM and Google place API\footnote{\url{https://developers.google.com/places/web-service/search}}.
\item Mobility Report ($MR$): The individual and aggregated mobility information are analysed for predicting hotspots. The aggregated mobility report is collected from \textit{Google Community mobility Report}{\footnote{\url{https://www.google.com/covid19/mobility/}}}. The individual mobility history is collected from users' smart-device and Google map timeline of individuals\footnote{\url{https://www.google.com/maps/timeline?pb}} in the study-region.
    \end{enumerate}

All of these information is managed using an efficient storage method. The region is segmented into different grids, and the fog devices store the information of all the grids within its coverage using a hashing scheme \cite{mario}. Thus, the information is segregated into different hierarchical segments, and data extraction becomes efficient. 
\par Next, it is important to find out the spatio-temporal correlation (SC) factor of the COVID-19 spread in neighbouring regions, which has been computed based on Moran's I \cite{moran1950notes} as follows:
\begin{equation}
    SC=\frac{o}{\sum_a \sum_b w_{ab}}\times \frac{\sum_a \sum_b w_{ab} (v_a - \Bar{v})(v_b - \Bar{v})}{\sum_a (v_b - \Bar{v})^2}
\end{equation}
 Here, $v_a$ represents the number of events/ incidence (here, number of new cases reported in region $a$) observed in a region $a$, the mean of the newly confirmed cases in the whole region is $\Bar{v}$, the total number of observations is $o$. $w_{ab}$ represents the spatial adjacency between $a$ and $b$. Here, we have used different types of adjacency relation instead of only distance measure. The direct spatial distance measures as follows: \textit{(1) two regions share a common border; (2)  direct route is present to reach $b$ directly from $a$}. Next, we sort the regions based on their \textit{(3) population density, (4) literacy rate, (5) available medical facilities, (6) aggregate movement flow}. Based on each of the variable, we compute $SC$ where a region is adjacent with its previous ranked region and the next ranked region. We compute these values at different time-spans (specifically in each week of the study period). 
 \begin{equation}
     SC=\left\{\begin{matrix}
0 & No \; Spatial \; Autocorrelation\\ 
>0 & Positive \; Spatial \; Autocorrelation\\ 
<0 & Negative \; Spatial \; Autocorrelation\\  
\end{matrix}\right.
\label{eqmoran}
 \end{equation}
 As shown in the equation (\ref{eqmoran}), when the computed value of SC is 0, it represents no spatial
autocorrelation. Similarly, the larger absolute SC value demonstrates stronger spatial autocorrelation. This information (SC on six different adjacency metric) is used in the deep learning module to efficiently predict next possible hotspot zones.
 
      \par Specifically, a trajectory segment/ movement history of an individual records the consecutive stay-points in the path of the user, which are continuous spatial locations. Each grid is identified with unique id. Thus, we discretize the continuous spatial location information. We use the skip-gram model to learn the representation of spatial locations ($l_1, l_2, \dots, l_T$) in the training dataset:
      \begin{equation}
          \frac{1}{T} \sum_{t=1}^T \log [p(l_{t-c}, \dots, l_{t-1}, l_{t+1}, \dots, l_{t+c}|l_t)]
      \end{equation}
      which can be written as:
      \begin{equation}
      \begin{split}
          = \frac{1}{T} \sum_{t=1} \sum_{con} \log p(l_{t+j}|l_t) \\
          where \; context: -c\leq j \leq c, j \neq 0
      \end{split}
                \end{equation}
                where $l_{t+j}$ is the neighboring location of the present location $l_t$. Based on the spatial proximity rule that two neighboring locations have similar representation, we use softmax function as defined:
                \begin{equation}
                   p(l_{t+j}|l_t) = \frac{\exp(a_{l_{t+j}}^N a_{l_t})}{\sum_{l_i \in l} \exp(a_{l_{i}}^N a_{l_t})}
                \end{equation}
                Thus, the embedding layer converts the spatial locations into: 
                \begin{equation}
                    a_{l_1}, a_{l_2} \dots, a_{l_N}
                \end{equation}
                where $N$ is the number of the regions.
                Next, we embed the temporal information. The representation of the temporal information should comprises of day of the week, timestamp and time-duration spent in a location. Here, the skip-gram model is not efficient, and we use paragraph-vector model and get the vector representation of temporal information. 
                After embedding the spatial and temporal features, we deploy a bidirectional LSTM layer to capture the shared information of both the tasks. The basic building block of the LSTM layer is as follows:
               \begin{equation}
    \begin{split}
        i_t= \sigma (W_i \cdot [h_{t-1},x_t]+b_i), \; f_t= \sigma (W_f^\cdot [h_{t-1},x_t]+b_f)\\
o_t= \sigma (W_o \cdot [h_{t-1},x_t]+b_o); \; \\
c_t=f_t\cdot c_{t-1} + i_t \cdot \tanh(W_c \cdot [h_{t-1},x_t]+b_c) 
    \end{split}
\end{equation}

 where the input, output, and forget gates are represented by $i,o$ and $f$, respectively. The hidden representation is denoted by $h_t$. Since, \textit{STOPPAGE} uses a
bidirectional LSTM, the output of the layer is modified as:
\begin{equation}
    h_t= \vv{h_t} + \overleftarrow{h_t} 
\end{equation}
Here, the output from the forward and backward propagation layer are represented by $\vv{h_t}$ and $\overleftarrow{h_t}$ respectively. 

\par The next layer utilizes a \textit{Gated recurrent unit (GRU)} which is similar to $LSTM$. This layer is used for extracting the impact of the other contexts such as demography, age and aggregated movement etc. GRU has reset and update gate which are formally defined as:
       \begin{equation}
           \begin{split}
               z_t^{'}=\sigma(W_z^{'} \cdot [h_{t-1}^{'},x_t^{'}]) \; \;            r_t^{'}=\sigma(W_r^{'} \cdot [h_{t-1}^{'},x_t^{'}]) \\
               h_t^{'}=(1-z_t^{'}) \ast h_{t-1}^{'}+z_t^{'} \ast \tanh(W_h^{'} \cdot [r_t^{'} \ast h_{t-1}^{'},x_t^{'}]
                         \end{split}
       \end{equation}
       where $W_z^{'}$, $W_r^{'}$ and $W_h^{'}$ are weight matrices. The \textit{update gate} ($z_t^{'}$) helps to extract the required information from the past time-step and pass to the future. On the other hand, \textit{reset gate} ($r_t^{'}$) is used to determine how much past information needs to be eliminated. Therefore, in the proposed model, the GRU is capable to filter and store information utilizing the reset and forget gates. Here, we use specific information such as $SC$, $P_{ca}$, $P_{co}$ etc. This is crucial to capture the different phase of the disease spread, since GRU layer eliminates the vanishing gradient problem, but pass the relevant information to the next steps of the network. Note that in the proposed approach, deep LSTM architecture is used, allowing the network to learn at different time scales over the input. Furthermore, they can make better use of parameters by distributing over the space through multiple layers. 
       \par Another important module of the architecture is \textit{attention mechanism}. It is used to capture the relationship between all of these contextual factors and hotspots in different spatio-temporal resolutions. Here, we have used the dot product attention function $f_{att}$ and the representation is defined as:
       \begin{equation}
           r^{att} = \sum_{i=1}^N \frac{\exp(f_{att}(h_{t_i}^{'},m_{t_i}))}{\sum_{i=1}^N \exp(f_{att}(h_{t_i}^{'}, m_{t_i}))} h_{t_i}^{'}
       \end{equation}
       Here, $m$ is the vector representation of spatio-temporal features from the previous layer. The GRU input layer is replaced by the weighted representation ($r_{att}$).  
       On the other side, air travel connectivity index is utilized for the initial phase of disease spread. Here, \textit{attention layer} provides more \textit{attention} to the long-distance travels and connectivity between regions. For both the phases of disease spread, the final layer is softmax layer where all the outputs from different layers are fed. 
       \begin{equation}
           \begin{split}
              y^{'}=softmax(W^{'}h^{'} + b^{'}) 
           \end{split}
       \end{equation}
       The network is trained to minimize the cross-entropy loss as:
       \begin{equation}
           L(y,y')=- \sum_{i=1}^{TS} \sum_{j=1}^{Cl} y'^j_{i} \log(y^j_i)
       \end{equation}
       The number of training samples and number of classes are represented by $TS$ and $Cl$. We define 4 classes here: (i) $>20$ cases within 500m; (ii) $>50$ cases within 1km; (iii) $<5$ cases within 1km  and (iv) $<10$ cases within 2km. We define (i) and (ii) as hotspot zones. $y'$ is the ground truth and $y$ is the prediction probability. 
      
\par It may be noted that, in the $MR$ feature analysis, we also combine the information if any neighborhood region was a containment zone/ hotspot zone (high infection rate) in last 14 days. This information is stored in the respective fog devices of the region. Thus, the architecture encodes and learns different mobility semantics and other parameters at varied contexts and finds out whether the region is the next hotspot zone. 

\par 

\section{Applications of \textbf{STOPPAGE} }
In this section, we show how \textit{STOPPAGE} can be utilized to combat pandemic situation. We have discussed two use-cases, and how data analytics engine and SDI of \textit{STOPPAGE} can be utilized in such applications. An Android application along with the API-endpoints have been developed to retrieve and share information. We have also shown delay and power consumption to illustrate the feasibility of \textit{STOPPAGE} in terms of latency and energy consumption. 
\subsection{Use-Case: Find the Suspected Crowd} This is also observed that the risk of the infection spread depends on other factors also. For instance, the probability of infecting other people are much higher when the asymptomatic person is in the workplace or shopping mall compared to when he is in a less crowded park or playground. Also, the time-duration of contact with other person is also an important factor. Therefore, \textit{STOPPAGE} can also use the individual mobility history to find the suspected people from the PKG analysis. If any of the users is tested COVID19-positive, then the information is send to the nearby fog device and based on the PKG analysis, all other users are notified.  Alongside, using the reverse pruning technique, we can also find out the asymptotic person by analysing the mobility data. Analysing and assisting people in real-time is the major challenge here. The fog devices keep track of a particular locality and assist users within that region. Since, the dataset is huge, any compute-intensive task is carried out in the distant cloud servers. However, since the fog nodes store the results, the delay in communication is reduced. 
\subsection{Use-Case: Health Status Monitoring and Assistance at Home}
In this crucial time, it may not be possible to clinically test everyone. However, the preliminary testing can be done at home. The basic health parameter values such as body temperature, blood pressure (systolic and diastolic), pulse rate, SPO2 (oxygen saturation) level can be collected using BAN and sent to the smart phone of the user. In case of COVID-19, the common symptoms are fever, abnormality in breathing, cough etc. Here, fever and breathing abnormality can be predicted from the collected body temperature, pulse rate and SPO2 level. The collected health parameter values are sent to the smart phone. The smart phones already have GPS (Global Positioning System) tracking module, the environmental temperature and humidity detecting module. It can check that whether the collected data values fall in the normal range with respect to the user's health profile and contextual information (location, motion and acceleration, environmental temperature and humidity). If the collected value falls in the normal range, the smart phone will show that the health status is normal.
\begin{figure}
    \centering
    \includegraphics[scale=0.80]{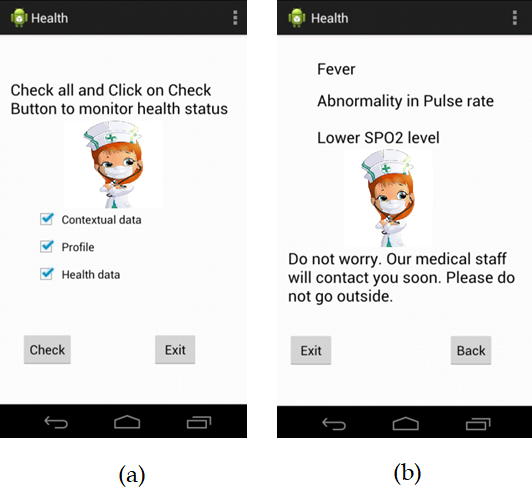}
    \caption{MyHealth: Android app for health monitoring}
    \label{myhe}
\end{figure}
Let the normal range of a health parameter $hp$ is $hp_{up_u}$ to $hp_{low_u}$, i.e. the upper value is $hp_{up_u}$ and lower value is $hp_{low_u}$ for a user $u$ with respect his/her health profile and contextual information, where $hp \in H$ and $H$ is the set of health parameters. Let $hp_{cl_u}$ is the collected value of the health parameter $hp$ for the user $u$. If $hp_{up_u} \geq hp_{cl_u} \geq hp_{low_u}$, then the collected health parameter value resides in normal range. This checking is done for all health parameters considered in the experiment. If $hp_{cl_u}$ falls in normal range for $hp$ $\forall H$, then the status is predicted as normal. Otherwise, the predicted health status is abnormal and the smart phone will show an alert message, and send the collected data with user's health profile and contextual information to the cloud. The health care centre of the region where the user belongs will access the data and contact the user for further check up. In this way a preliminary health monitoring service can be provided to the user. 


\subsubsection{MyHealth: An Android App for Health Monitoring}
We have designed an Android application (app) called, MyHealth which can be used for personal health status checking. The collected health data, profile and contextual information are processed, and if the user's health status seems to be abnormal (e.g. body temperature is high, pulse rate is abnormal, SPO2 level is low), then the user is suggested to be at home, and medical person will contact him/her soon for further medical tests (refer Fig. \ref{myhe}). It indicates faster health care provisioning than the traditional system, where the user has to contact a medical staff by himself/herself for further medical tests all the time. 
\subsubsection{Delay in Health Status Prediction}
\begin{figure}
    \centering
    \includegraphics[scale=0.4]{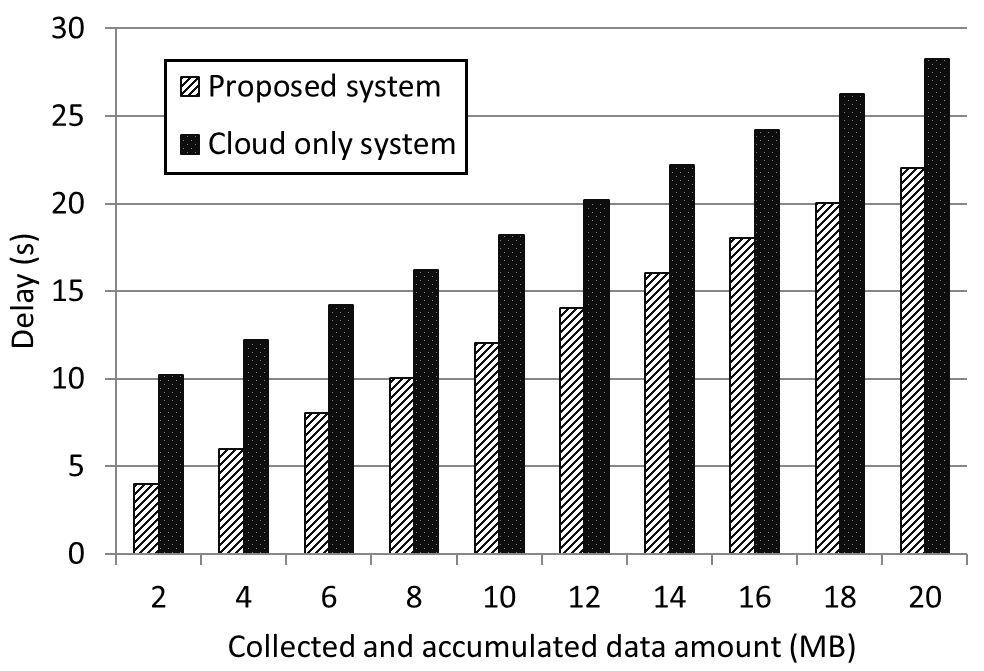}    
    \caption{Delay in health status prediction}
    \label{del}
\end{figure}

Let the total delay in collecting health parameter data and transmission to the smart phone is $Delay_h$, and the delay in collecting contextual information is $Delay_c$. The health profile is already present inside the smart phone. As the contextual information collection takes place simultaneously while the smart phone receives health parameter data from BAN, the total delay in collecting and accumulating health data with profile and contextual information is given as:
\begin{equation}
\label{e1}
    Delay_{ca}=Delay_{mob}+max(Delay_h, Delay_c)
\end{equation}
where $Delay_{mob}$ is the delay in data accumulation.
Let the amount of data transmitted in uplink communication $i$ is $D_i$, and in downlink communication $j$ is $D_j$. Let the data transmission rate in uplink communication $i$ is $R_{ui}$, in downlink communication $j$ is $R_{dj}$. Let the failure rate in uplink communication $i$ is $f_{ui}$, in downlink communication is $f_{dj}$ respectively.  If there are $m$ uplink communications and $n$ downlink communications take place, then the total communication delay is given as,
\begin{equation}
\label{e2}
\begin{split}
    Delay_{com}=\sum_{i=1}^{m}{(1+f_{ui})\cdot(D_i/R_{ui})}+\\
    \sum_{j=1}^{n}{(1+f_{dj})\cdot(D_j/R_{dj})}
\end{split}
\end{equation}
Let the amount of data processed inside the smart phone is $D_{mob}$, inside the fog device is $D_f$, and inside the cloud is $D_c$, and the processing speed of the smart phone, fog device and cloud are $S_{mob}$, $S_f$ and $S_c$ respectively. Then the delay in processing the data is given as,
\begin{equation}
 \label{e3}
    Delay_{pro}=(D_{mob}/S_{mob})+(D_{f}/S_{f})+(D_{c}/S_{c})
\end{equation}
The total delay in predicting health status is given as,
\begin{equation}
\label{e4}
    Delay_{tot}=Delay_{ca}+Delay_{com}+Delay_{pro}
\end{equation}
The delay in predicting health status in case of proposed system and in case of cloud only system are presented in Fig. \ref{del}. This is observed that the use of the proposed health care system reduces the delay $\sim$(21-60)\% than the cloud only health care system. In cloud only system the communication delay is higher, where as in proposed system the intermediate nodes participate in data processing, which in turn reduces the communication delay. Subsequently, the total delay is reduced in the proposed system.
\begin{figure}
    \centering
    \includegraphics[scale=0.43]{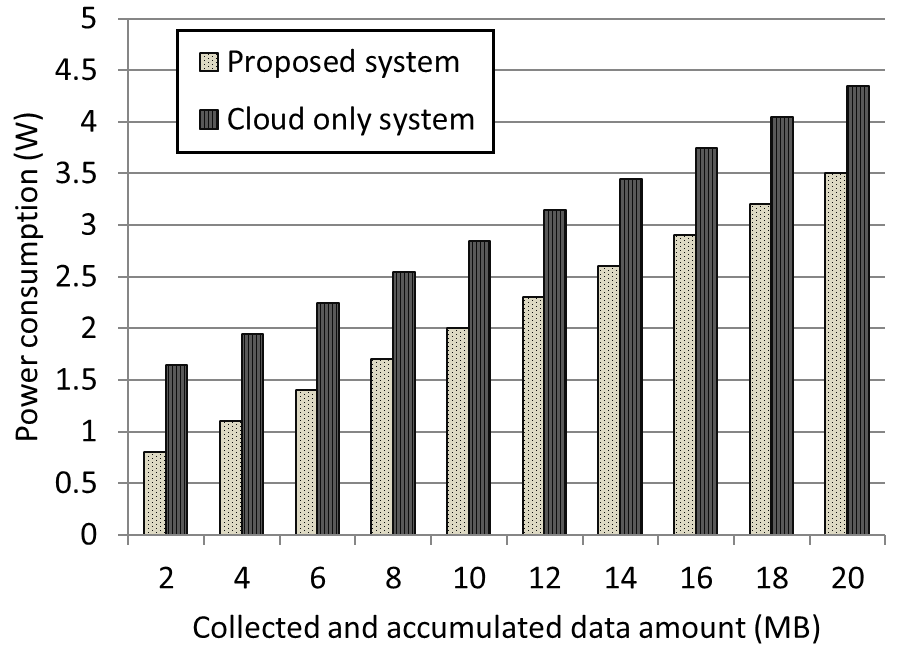}    
    \caption{Power consumption of the smart phone during health status prediction period}
    \label{powr}
\end{figure}
\subsubsection{Power Consumption of Smartphone during Health Status Prediction/ Monitoring} 
 In the module, smartphones are used as an edge device, which collects, accumulates and sends health and context related information. In this context, it is important to measure and optimize the power consumption of the mobile device. Let the power consumption of the smart phone in data transmission, reception, active and idle mode are $P_t$, $P_r$, $P_a$, and $P_i$ respectively. 
 
 \begin{figure*}
    \centering
    \includegraphics[scale=0.75]{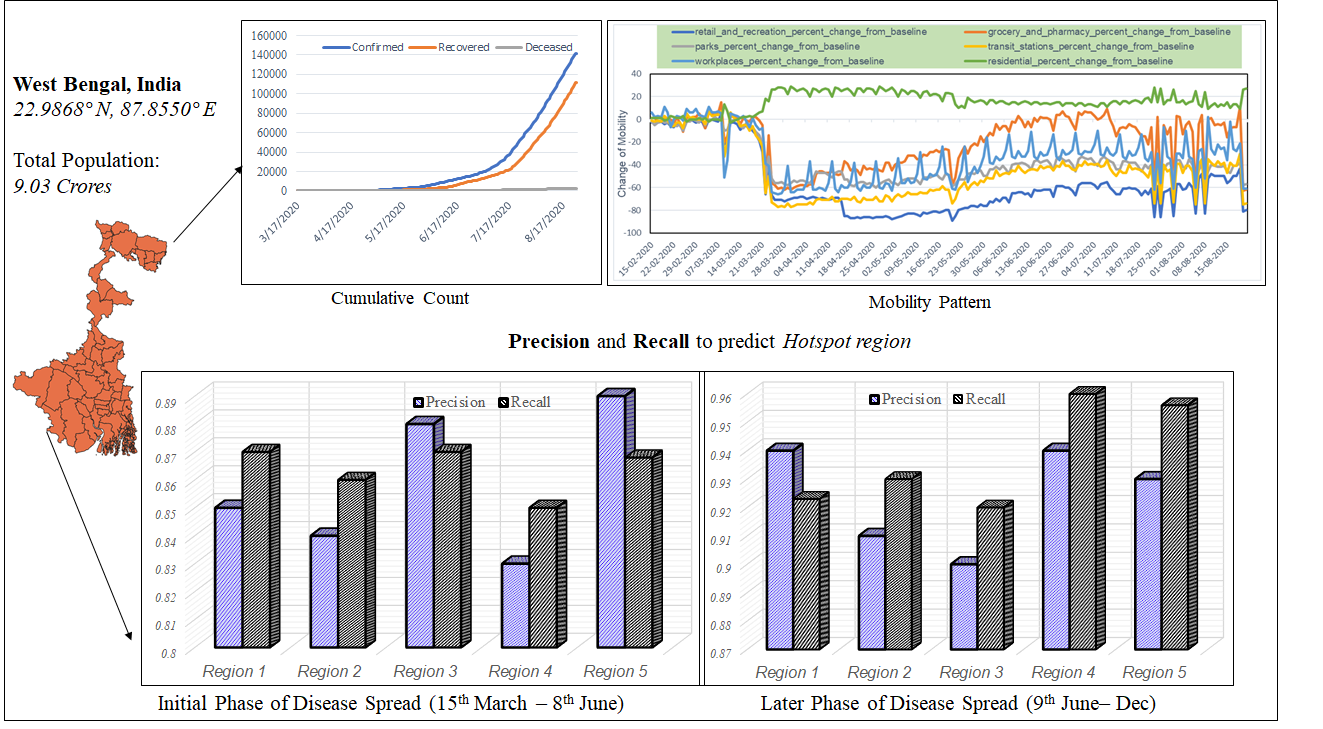}
    \caption{Experimental Result \textit{(Five regions have been selected and the boundary of all hotspots within the regions have been predicted by \textbf{STOPPAGE})}}
    \label{fig:exp1}
\end{figure*}

The power consumption of the smart phone during data collection and accumulation period is given as,
\begin{equation}
\label{e5}
    P_{ca}=(P_a\cdot Delay_{mob})+(P_r\cdot max(Delay_h, Delay_c))
\end{equation}
Let for the smart phone the number of uplink communication is $k$ and downlink communication is $q$. Then the power consumption of the smart phone during communication period is given as,

\begin{equation}
    \label{e6}
    \begin{split}
        P_{com}=(P_t\cdot(\sum_{i=1}^{k}{(1+f_{ui})\cdot(D_i/R_{ui})}))+\\
    (P_r\cdot(\sum_{j=1}^{q}{(1+f_{dj})\cdot(D_j/R_{dj})}))+\\
    (P_i\cdot(\sum_{i=1}^{(m-k)}{(1+f_{ui})\cdot(D_i/R_{ui})}))+\\
    (P_i\cdot(\sum_{j=1}^{(n-q)}{(1+f_{dj})\cdot(D_j/R_{dj})}))
    \end{split}
\end{equation}
The power consumption of the smart phone during data processing period is given as,
\begin{equation}
 \label{e7}
    P_{pro}=P_a\cdot(D_{mob}/S_{mob})+P_i\cdot(D_{f}/S_{f})+P_i\cdot(D_{c}/S_{c})
\end{equation}
The power consumption of the smart phone in the proposed system and cloud only system are presented in Fig. \ref{powr}. As in the proposed system the intermediate nodes participate in data processing the delay is reduced, and consequently the power consumption of the smart phone is also reduced. This is observed that the proposed system reduces the power consumption of the smart phone by $\sim$(19-50)\% than the cloud only system.
\begin{figure}
    \centering
    \includegraphics[scale=0.49]{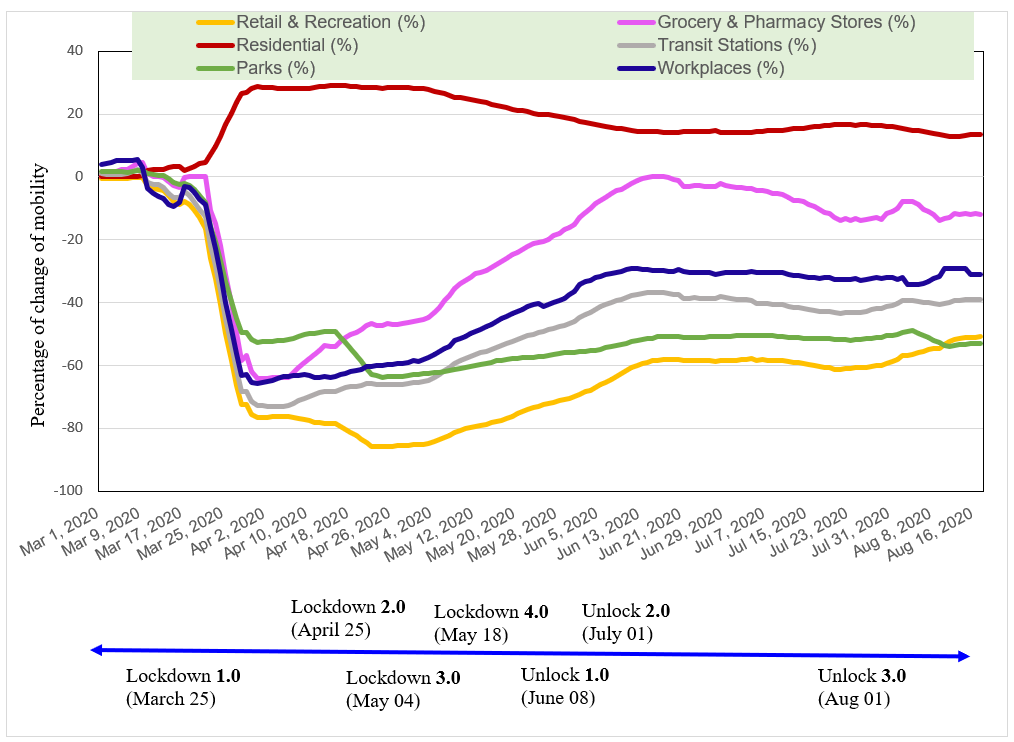}
    \caption{Mobility pattern change across India}
    \label{overallmobi}
\end{figure}

\par In the conventional strategies followed in most of the developing countries, the user sees the parameter values, answers to the questions regarding health monitoring over voice call/message, and then based on his/her answers, the health status is predicted, and if it seems to be abnormal, then the user is asked to visit a health centre for further check up. It is quite obvious this whole procedure takes much time. If the mobile health monitoring takes place, the latency is quite less than the conventional health monitoring system.  Moreover, as the data collection and processing entirely takes place by the device, the accuracy of the prediction and reliability are also higher. On the other hand, the user can do this residing at home; no need to visit the health centre for preliminary check up until and unless an emergency occurs. This in turn helps to avoid unnecessary gathering and social distancing. Thus, we strongly recommend to use mobile health monitoring in such situations to avoid unnecessary delay and gathering.

\begin{table*}[h]
\centering
 \caption{Comparison of Accuracy Measure and Ablation Study of the Deep Learning Module for Identifying COVID-19 Hotspots}
 \begin{tabular}{|c|c|c|c|} 
 \hline \hline 
 \textit{Model}  & Precision & Recall & Accuracy \\ \hline 
 KNN & 54.03\% & 54.8\% & 56.05\% \\ \hline  
 DT & 50.08\% & 51.87\% & 54.9\% \\ \hline  
 SVM & 60.04\% & 58.56\% & 62.48\% \\ \hline  
 CNN & 75.12\% & 72.30\% & 77.15\%  \\ \hline
 ST-RNN & 77.05\% & 72.68\% & 78.72\%  \\ \hline  \multicolumn{4}{|c|}{\textbf{\textit{\hspace{0.3cm} Ablation Study of the Deep Learning Module of \textbf{STOPPAGE}}}} \\
\noalign{
\color{black}
\hrule height 2pt
}%
w/o PKG Integration & 82.8\% & 83.74\% & 86.18\%   \\ \hline 
w/o Attention Layer   & 84.4\% & 85.78\% & 88.09\%   \\ \hline 
w/o bi-LSTM   & 85.20\% & 87.51\% & 93.01\% \\ \hline 
w/o two-phase Analysis & 82.04\% & 86.18\% & 90.45\%  \\ \hline 
 \textbf{STOPPAGE} (Full) &  \textbf{88.15}\% & \textbf{90.25}\% & \textbf{93.12}\% \\ \hline 

 \end{tabular}\label{expact}
 \end{table*}
\section{Implementation and Evaluation}
In this section, we present the implementation details of the proposed framework, \textit{STOPPAGE}. We have used the compute engine and app engine of Google Cloud Platform (GCP) to carry out the spatio-temporal data analysis. In the test-bed, we have used  Raspberry Pi 3 as the fog device. We have also designed an Android application, using Android Studio 4.1 with \textit{Firebase database} support, which collects the location, acceleration, proximity, temperature and light sensor data from the smartphone's in-built sensors using the \textit{Android sensor framework}. The application is also capable to communicate with the wearable devices such as smart-watch (Fitbit). In the Raspberry Pi 3, we have installed the \textit{Eddystone Bluetooth Beacon}, for sending data periodically.

\begin{table}[!htb]
           \caption{Comparison of Execution time of \textbf{STOPPAGE} for retrieving PKG information}
           \label{exect}
    \resizebox{0.45\textwidth}{!}{%
    \begin{tabular}{|l|l|l|l|l} \hline 
             No. of Entities &  \multicolumn{3}{c|}{Execution Time (in minute)}\\ \cline{2-4}
             & Huang et al. \cite{baselinekg} &  Trivedi et al. \cite{tkg} & \textbf{STOPPAGE}   \\ \hline
              $1 \times 10^3$ & 2.2 & 2.5 & 2.6 \\ \hline
            $5 \times 10^3$  & 8.9 & 7.4 & 3.23 \\ \hline
             $10 \times 10^3$  & 26.01 & 20.4 & 5.2 \\ \hline
              $20 \times 10^3$ & 35.18 & 28.7 & 9.12 \\ \hline
              $40 \times 10^3$ & 40.54 & 34.2 & 16.5 \\ \hline
             $50 \times 10^3$ & 49.2 & 40.4 & 21.32 \\ \hline
                                                    \end{tabular}%
                                 }
                                 \end{table}

\par For training the deep learning module, we have used popular Adam algorithm to update the network weights iteratively and optimize the parameters. The run-time (training time) of the model is 86minutes for all five regions for 100 rounds (epochs). We also observe the performance (accuracy@5 for hotspot prediction) variation of the model using different cell size and batch size, and get optimal results in cell size $64$ and batch size $10$.

\par For evaluating the spatio-temporal analysis, we have implemented the methods in GCP VM and QGIS framework. The overall mobility change in India is shown in Fig. \ref{overallmobi}. It also shows the phase-wise lockdown and unlockdown measures taken by the government. Based on the overall situation of the disease spread, we have evaluated the prediction in two phases: (i) initial phase of the disease spread (upto unlock 1.0) and (ii) next phase (till mid Dec). The accuracy of predicting the hotspots is shown in Fig. \ref{fig:exp1}. For this prediction, we have considered the study region as West Bengal, a state of India. We have executed the proposed method on five selected regions of the state where we have the access of individual mobility history of the residents. The precision and recall value computation captures the efficacy of \textit{STOPPAGE} to predict the boundary of the hotspot region. The ground-truth data has been collected from the state government dashboard\footnote{\url{https://wb.gov.in/containment-zones-in-west-bengal.aspx}} in daily basis. It is observed that the precision and recall in both phases of the disease spread is quite high. \textit{STOPPAGE} can identify the hotspot areas apriori with more than $85\%$ precision and recall values. Table \ref{expact} presents the accuracy measure of \textit{STOPPAGE} along with the baseline methods namely, \textit{KNN}, \textit{Decision Tree (DT)}, \textit{SVM}, \textit{CNN}\cite{cnn} and \textit{ST-RNN}\cite{strnn}. The parameter for KNN is selected as $5$. We have chosen \textit{radial basis function (RF)} as the activation function in NN. A linear kernel is selected for SVM. The results for different runs are captured and average precision, recall and accuracy measure is reported. We have performed the ablation study of the deep learning architecture of \textit{STOPPAGE} to demonstrate the significance of each of the modules. Table \ref{expact} shows the comparison values of the baselines and the ablation result. It is observed that \textit{STOPPAGE} outperforms all other baselines in a significant margin. The ablation study proves that the architecture is suitable and better compared to all other possible set-ups. This shows the novelty and the significance of the deep learning module of \textit{STOPPAGE} in COVID-19 context. The Android application can be used to monitor the current pandemic status of the country, and for assisting the users. It retrieves the data from the nearest fog device regarding the predicted hotspot areas and sends the notification to the users within the predicted hotspots. 

\begin{figure}
    \centering
    \includegraphics[scale=0.70]{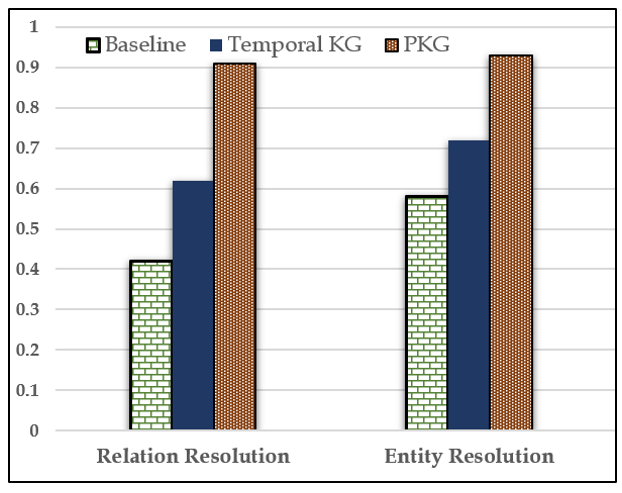}
    \caption{Accuracy of PKG}
    \label{exp3}
\end{figure}

\par One of the major contributions of this paper is constructing the Pandemic-Knowledge graph (PKG) and summarizing the information in the knowledge graph structure. We have implemented PKG with real-life dataset and compared the execution time to extract relations or facts with different number of entities ranging from 1000 to 50,000. The performance of PKG is compared with baseline knowledge graph \cite{baselinekg} and temporal knowledge graph \cite{tkg}. The execution time to extract the facts or relations is shown in TABLE \ref{exect}. It has been observed that our proposed framework \textit{STOPPAGE} has lower execution time than the other existing approaches, which indicates \textit{STOPPAGE} provides better time-efficiency. Fig. \ref{exp3} shows the accuracy measures for two tasks namely, \textit{relation resolution} and \textit{entity resolution}. The \textit{relation resolution} finds out facts or relations from the knowledge graph, and \textit{entity resolution} refines the entity information effectively. In the temporal knowledge graph \cite{tkg}, the authors present a novel deep learning based framework to combine the evolving entities and their dynamically changing relationships over time, and represented that as \textit{knowledge evolution}. However, in the context of COVID-19, this approach does not work well. It is observed that the baseline and temporal KG have achieved accuracy within $0.42-0.72$ range, while \textit{STOPPAGE} has an accuracy of $0.91-0.93$ range. Therefore, it is observed empirically that \textit{STOPPAGE} has outperformed other existing approaches in a larger margin in terms of accuracy and execution time.

\section{Conclusions and Future work}
In this paper, we have presented an efficient pandemic management and monitoring framework, namely, \textit{STOPPAGE} leveraging the Internet of Spatio-Health Things - a Cloud/Fog/Edge/IoT based efficient SDI to overcome the various present challenges due to the outbreak of COVID-19. Specifically, we have presented a novel spatio-temporal data analytics method using deep learning to augment varied contextual information to identify probable hotspots for taking preventive measures. A novel pandemic knowledge graph (PKG) is proposed to find out the correlations of disease outbreak and semantic spatio-temporal information. \textit{STOPPAGE} presents varied types of movement semantics (cascading and co-occurrence patterns), and spatio-temporal correlation with the COVID-19 spread which are potent features to predict next hotspot zones. 
The proposed fog/edge based architecture outperforms existing approaches in terms of delay, power consumption and accuracy. It is shown that SDI is one of the major components to combat the pandemic situation. In the future, we will explore the possibility of incorporating climatology data to enhance the accuracy of \textit{STOPPAGE}. \textit{STOPPAGE} facilitates a spatio-temporal data-driven analytics engine running over SDI backbone which is capable to extract meaningful insights from available data utilizing technologies like IoT, mobility analytics in spatial context. However, in future, we will extend \textit{STOPPAGE} by incorporating domain knowledge from medical and epidemiology experts to make the system more robust and reliable to combat any type of pandemic situation efficiently.  
\section*{Acknowledgment}
The work is partially supported by DST Research Grant. The authors are thankful to the local medical authority and stakeholders for supporting this work.

\bibliography{biblio}
\end{document}